\documentclass[journal]{IEEEtran}
\usepackage{amsmath,amsfonts}
\usepackage{algorithmic}
\usepackage{algorithm}
\usepackage{array}
\usepackage[caption=false,font=normalsize,labelfont=sf,textfont=sf]{subfig}
\usepackage{textcomp}
\usepackage{stfloats}
\usepackage{url}
\usepackage{verbatim}
\usepackage{graphicx}
\usepackage{cite}
\usepackage{color}
\usepackage{pifont}
\usepackage{multirow}
\usepackage{booktabs}
\usepackage{makecell}
\begin{document}

\title{GAA-TSO: Geometry-Aware Assisted Depth Completion for Transparent and Specular Objects}

\author{Yizhe Liu, Tong Jia, Da Cai, Hao Wang, Dongyue Chen
\thanks{This work was supported in part by the National Key Research and Development Project of China under Grant 2022YFF0902401, in part by the National Natural Science Foundation of China under Grant U22A2063 and 62173083, in part by the Major Program of National Natural Science Foundation of China under Grant 71790614, in part by the 111 Project under Grant B16009, and in part by the Liaoning Provincial "Selecting the Best Candidates by Opening Competition Mechanism" Science and Technology Program under Grant 2023JH1/10400045. \textit{(Corresponding author: Tong Jia.)}}
\thanks{Yizhe Liu, Da Cai, Hao Wang, Dongyue Chen are with the College of Information Science and Engineering, Northeastern University, Shenyang 110819, China (e-mail: 2110335@stu.neu.edu.cn; 2190040@stu.neu.edu.cn; wanghao@ise.neu.edu.cn; chendongyue@ise.neu.edu.cn).}
\thanks{Tong Jia is with the College of Information Science and Engineering, Northeastern University, Shenyang 110819, China, and with the National Frontiers Science Center for Industrial Intelligence and Systems Optimization, Shenyang 110819, China (e-mail: jiatong@ise.neu.edu.cn).}}



\maketitle

\begin{abstract}
Transparent and specular objects are frequently encountered in daily life, factories, and laboratories. However, due to the unique optical properties, the depth information on these objects is usually incomplete and inaccurate, which poses significant challenges for downstream robotics tasks. Therefore, it is crucial to accurately restore the depth information of transparent and specular objects. Previous depth completion methods for these objects usually use RGB information as an additional channel of the depth image to perform depth prediction. Due to the poor-texture characteristics of transparent and specular objects, these methods that rely heavily on color information tend to generate structure-less depth predictions. Moreover, these 2D methods cannot effectively explore the 3D structure hidden in the depth channel, resulting in depth ambiguity. To this end, we propose a geometry-aware assisted depth completion method for transparent and specular objects, which focuses on exploring the 3D structural cues of the scene. Specifically, besides extracting 2D features from RGB-D input, we back-project the input depth to a point cloud and build the 3D branch to extract hierarchical scene-level 3D structural features. To exploit 3D geometric information, we design several gated cross-modal fusion modules to effectively propagate multi-level 3D geometric features to the image branch. In addition, we propose an adaptive correlation aggregation strategy to appropriately assign 3D features to the corresponding 2D features. Extensive experiments on ClearGrasp, OOD, TransCG, and STD datasets show that our method outperforms other state-of-the-art methods. We further demonstrate that our method significantly enhances the performance of downstream robotic grasping tasks.
\end{abstract}


\section{Introduction}
\IEEEPARstart{T}{ransparent} and specular objects are widely used in daily life, chemical laboratories, and manufacturing. However, due to their unique reflection and refraction properties, depth cameras such as time-of-flight \cite{frangez2022assessment} or structured light \cite{liu2023joastereo} often generate inaccurate and missing depth information. This challenge seriously hinders robots from performing depth-based intelligent grasping and manipulation of these objects. Therefore, researching methods that can accurately restore the depth values of these objects is particularly important in practice.

\begin{figure}[!t]
\centering
\includegraphics[width=1. \linewidth]{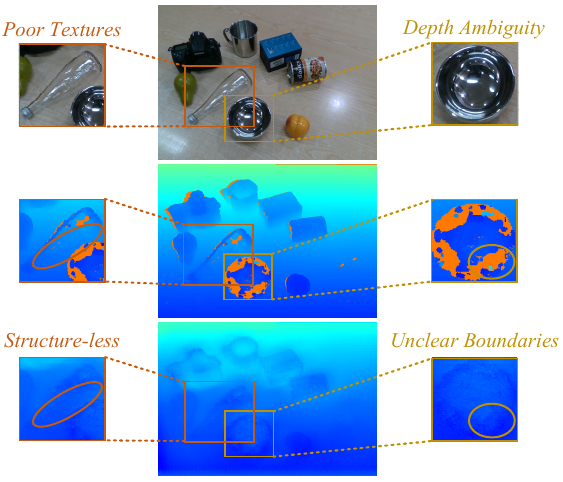}
\caption{An example of depth completion in a scene containing transparent and specular objects. From the first to the third row, we show the RGB image, the raw depth map, and the depth completion result from FDCT \cite{li2023fdct} respectively.}
\label{introduction_show}
\end{figure}

Recent depth completion methods for transparent and specular objects usually leverage the aligned RGB image as a guidance, assuming that RGB information can provide rich color and scene structure (e.g., textures) cues for missing pixels. Specifically, these methods \cite{sajjan2020clear, dai2022domain} utilize the raw RGB and depth image pairs from the depth sensor to restore the depth map. However, transparent and specular objects often appear as poor-texture regions in RGB images, making networks that heavily rely on color information tend to give roughly the same guidance to these regions, which leads to structure-less depth predictions, as shown in the left box of Fig. \ref{introduction_show}.

Although some works \cite{zhu2021rgb, fang2022transcg, li2023fdct, zhai2024tcrnet} explore geometric constraints in depth maps by introducing surface normal, applying 2D convolutions on irregularly distributed depth values may cause implicit and ineffective exploration of the underlying 3D geometric structure, resulting in ambiguous depth predictions with unclear boundaries, as shown in the right box of Fig. \ref{introduction_show}. Therefore, considering the potential disadvantages of poor-texture and depth ambiguity, previous methods that only process RGB-D inputs at the image level have great limitations for the depth completion task of transparent and specular objects.

To address these issues, we are interested in how to explore the implicit geometric structure in depth information and utilize it to complement the 2D appearance features to recover more accurate depth maps. In order to effectively extract geometric structures, we design an auxiliary 3D branch to explicitly learn scene-level layout and object-level shape information. Combining this structure knowledge, our depth completion method can provide additional geometric cues for transparent and specular objects. 

Specifically, we propose a geometry-aware assisted depth completion method for transparent and specular objects called GAA-TSO, which includes three components: image branch, point cloud branch, and gated cross-modal fusion module. First, we use the image branch to acquire image features from the RGB-D input. For the point cloud branch, we back-project the raw depth of the input to a point cloud, which is then fed into a point cloud completion network to learn the geometric structure and obtain point-level 3D features. To effectively utilize 3D geometric features, we design several gated cross-modal fusion modules, which can propagate the captured hierarchical 3D structures to the image branch. In addition, we propose an adaptive correlation aggregation strategy to alleviate the projection misalignment problem caused by depth map error. Finally, multiple cascaded hourglass networks in the image branch collect multi-scale fusion features and output the final depth prediction. Extensive experiments show that our method achieves superior performance on ClearGrasp \cite{sajjan2020clear}, OOD \cite{zhu2021rgb}, TransCG \cite{fang2022transcg}, and STD \cite{dai2022domain} datasets. In summary, the main contributions of this article are as follows:

\begin{itemize}
    \item 
    We propose a novel geometry-aware assisted depth completion architecture for transparent and specular objects to explicitly explore and utilize 3D geometric structure information.
\end{itemize}

\begin{itemize}
    \item 
    We design several gated cross-modal fusion modules to adaptively propagate relevant 3D geometric features into 2D features and autonomously select useful cross-modal features through the gating mechanism.
\end{itemize}

\begin{itemize}
    \item 
    We propose an adaptive correlation aggregation strategy to alleviate the projection misalignment problem in cross-modal interaction, which enables efficient and accurate feature matching.
\end{itemize}

\begin{itemize}
    \item 
    Our method achieves superior performance on ClearGrasp, OOD, TransCG, and STD datasets. In addition, real-world experiments also show that our GAA-TSO benefits downstream robotic grasping tasks.
\end{itemize}

\section{Related Work}\label{related work}

\noindent \textbf{Depth Completion for Transparent and Specular Objects.}
As a fundamental task in computer vision, depth completion usually uses the raw depth and RGB images to restore the missing depth values in the depth map \cite{chen2023agg, zhou2023bev, wang2022rgb}. Due to the unique optical properties, depth maps captured by depth sensors on transparent and specular objects often contain inaccurate and missing depth values, which seriously affects downstream robotic tasks such as grasping \cite{fang2023anygrasp} and manipulation\cite{lu2024manigaussian}. To overcome this challenge, some researchers focus on depth completion of these objects based on the RGB-D input in recent years \cite{jiang2023robotic,yan2024transparent}. Sajjan et al. \cite{sajjan2020clear} presented a two-stage approach that first predicts surface normals, segmentation masks, and occlusion boundaries, followed by depth refinement through global optimization. Building on this, Tang et al. \cite{tang2021depthgrasp} employed a generative adversarial network to replace the computationally intensive global optimization step. Zhu et al. \cite{zhu2021rgb} developed a local implicit representation combined with the iterative self-correcting structure for depth completion. Xu et al. \cite{xu2021seeing} introduced a joint completion method for depth and point clouds to enhance the perception of transparent objects in complex environments. Fang et al. \cite{fang2022transcg} presented a lightweight deep network with the U-Net \cite{ronneberger2015u} architecture and utilized the GraspNet-baseline \cite{fang2020graspnet} for 6-DoF grasping pose detection. Similarly, Dai et al. \cite{dai2022domain} introduced a parallel Swin Transformer \cite{swinTransformer} RGB-D fusion network for depth restoration. To capture fine-grained features more effectively, Chen et al. \cite{chen2023tode} proposed a depth completion network utilizing a Transformer \cite{vaswani2017attention} encoder-decoder architecture. Li et al. \cite{li2023fdct} presented a fast depth completion framework capable of extracting both global and local features while maintaining low computational complexity. To maintain accuracy and real-time performance, Zhai et al. \cite{zhai2024tcrnet} introduced a depth completion method with a cascaded refinement structure for transparent objects. Sun et al. \cite{sun2024diffusion} introduced a diffusion-based depth completion architecture, which consists of two stages: region proposal and depth restoration. Different from existing methods, we propose a geometry-aware assisted depth completion method to fully explore the 3D geometric cues for transparent and specular objects.

\begin{figure*}[!t]
\centering
\includegraphics[width=0.98 \linewidth]{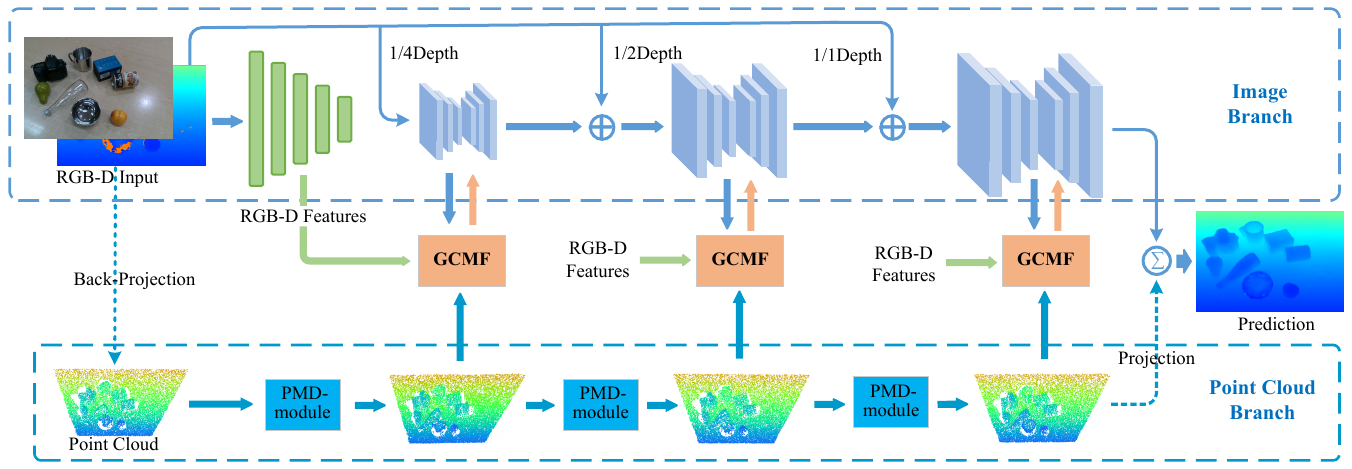}
\caption{The overview of our proposed GAA-TSO architecture, which consists of three main components: image branch, point cloud branch, and gated cross-modal fusion (GCMF).}
\label{architecture}
\end{figure*}

\noindent \textbf{2D-3D Cross-Modal Representation Learning.} Cross-modal representation learning is intended to capture shared feature representations from different types of modalities, which enables information from different modalities to complement and enhance each other \cite{song2023graphalign,li2023logonet}, making it widely explored in fields such as robotics, autonomous driving, and computer vision. Due to the limitations of 2D convolution in capturing 3D structural information, Chen et al. \cite{chen2019learning} designed a twin network to obtain 2D and 3D features respectively and fuse them in 2D image space. Qiu et al. \cite{qiu2019deeplidar} utilized surface normals as an intermediate bridge and further designed an enhanced encoder-decoder structure to efficiently integrate depth and color images. Similarly, Xu et al. \cite{xu2019depth} first simultaneously predicted coarse depth, confidence map, and surface normal from LiDAR inputs and then passed them to a diffusion refinement model to produce a denser depth map. Liu et al. \cite{liu20213d} proposed a distillation network to process 3D features into 2D, which introduced an additional 3D network to extract 3D features during the training phase to enhance the 2D features of the image branch. Xie et al. \cite{xie2023sparsefusion} used parallel camera and LiDAR detector outputs as fusion candidates and designed a lightweight self-attention module to fuse cross-modal candidates in 3D space. Liu et al. \cite{liu2023bevfusion} proposed an efficient and general multi-sensor fusion architecture that unifies multimodal information using a bird's-eye view representation to well preserve geometric and semantic information. In this article, we propose several gated cross-modal modules for transparent and specular objects to effectively fuse 2D and 3D features.

\noindent \textbf{Robotic Grasping.} With recent progress in deep learning, robotic grasping technology has experienced significant advancements \cite{newbury2023deep}. These grasping methods are divided into two categories: 4-DOF grasping methods \cite{qin2023rgb,yangGrasp,kumra2020antipodal} and 6-DOF grasping methods \cite{fang2020graspnet,breyer2021volumetric,chen2023keypoint,wu2024economic}. For 4-DOF grasping methods, Qin et al. \cite{qin2023rgb} designed an innovative grasp detection network that employs cross-modal attention to enhance the utilization of geometric information. Yang et al. \cite{yangGrasp}  presented a transformer-based grasping network MCT-Grasp, which improves grasping accuracy by learning the spatial relationship and geometric information of objects. For 6-DOF grasping methods, Fang et al. \cite{fang2020graspnet} introduced an end-to-end network for grasp pose estimation using point clouds as input, which decouples the learning of approach direction and operation parameters. Using 3D voxel representations of the scene as input, Breyer et al. \cite{breyer2021volumetric} designed a volumetric grasping network that simultaneously predicts grasp quality, opening width, and gripper orientation for each voxel. These grasping methods depend on precise depth information, making them unsuitable for grasping transparent and specular objects. To overcome this, a geometry-aware assisted depth completion method is designed to enhance the accuracy of robotic grasping for such objects.

\section{Proposed Approach}
\subsection{Overview}
Different from existing works, we focus on leveraging 3D geometry guidance to enhance depth completion for transparent and specular objects. To achieve this goal, we develop an auxiliary point cloud branch to learn 3D geometric features. The architecture of our GAA-TSO is shown in Fig. \ref{architecture}, which consists of an image branch, a point cloud branch, and several gated cross-modal fusion modules. Specifically, our image branch inputs raw depth and RGB images to extract multi-scale 2D features. To fully explore the 3D geometric information, we extract the depth channel from the input RGB-D and back-project it to the point cloud to learn 3D features. Since the raw depth map has missing depth values, we introduce a 3D completion network for the point cloud branch to gradually aggregate point-level 3D features and recover the complete scene. For the acquired 2D and 3D features, we designed the gated cross-modal fusion (GCMF) modules as the bridges to propagate 3D features to the 2D image branch, and combined with the 2D features through the gating mechanism to autonomously select appropriate fusion features. In addition, we propose an adaptive correlation aggregation strategy in GCMF to alleviate the projection misalignment problem caused by the depth map error. Finally, the fused features are fed into multiple cascaded hourglass networks in the image branch to refine the depth prediction.

\subsection{Image Branch}
The image branch aims to obtain 2D features and restore accurate depth maps with the assistance of 3D features. As illustrated in Fig. \ref{architecture}, our image branch includes a feature extractor and multiple cascaded hourglass networks. We concatenate the raw depth map with its corresponding RGB image and employ ResNet \cite{resnet} to extract multi-scale image features. Inspired by \cite{li2020multi}, multiple cascaded hourglass networks take depth maps of different resolutions as input and perform depth prediction at the same resolution. Each cascaded hourglass network is able to capture scene content within several layers and contains different levels of details. Specifically, the hourglass network with low-resolution input extracts the rough scene structures and gives a coarse depth prediction. The latter hourglass network uses the coarse depth prediction as a reference to generate a more fine-grained depth prediction. Each hourglass network receives the fused features of the corresponding scale from the gated cross-modal fusion module, which brings together 2D image features and 3D geometric features. In our architecture, the image branch consists of three cascaded hourglass networks that gradually refine to produce full-scale depth predictions. In addition, each stage generates a confidence map of the same size as the depth prediction to guide subsequent fusion. Therefore, the cascaded hourglass network outputs quarter, half, and full resolution respectively.

\subsection{Point Cloud Branch}
To capture and utilize geometric cues in the depth channel, we design a 3D branch to process the point cloud back-projected from the depth map. Given an input raw depth map $D$ of size $(H, W)$, we first obtain image coordinates $\mathcal{C}$, 
\begin{equation}
    \mathcal{C} = \{ (u, v, D(u, v)) | u \in [1, W], v \in [1, H] \}.
\end{equation}
Then we convert the image coordinates to the point cloud using the camera intrinsics $\mathcal{K} \in \mathbb{R}^{4 \times 4}$ and extrinsics $\mathcal{T} \in \mathbb{R}^{4 \times 4}$. For the $i$-th image coordinate $\mathcal{C}_i = (u_i, v_i, d_i)$, the corresponding 3D point $\mathcal{P}_i = (x_i, y_i, z_i)$ is:

\begin{equation}
    [x_i, y_i, z_i, 1]^T = \mathcal{T}^{-1} \cdot \mathcal{K}^{-1} \cdot [u_i \times d_i, v_i \times d_i, d_i, 1]^T.
    \label{2DTo3D}
\end{equation}

Since the input depth map has missing and inaccurate depth values, we adopt a 3D completion network to learn geometric features. Here, we introduce PMP-Net \cite{wen2021pmp}, which can capture the detailed topological and geometric structure relationship of the scene. Specifically, PMP-Net takes the back-projected point cloud as input, and gradually extracts point-level features and obtains the completed point cloud through three cascaded point moving distance (PMD) modules. The PMD-module in each step first acquires the global features from the input point cloud and propagates them to each point in the space through the feature propagation module, generating point-level 3D features. Afterwards, the 3D features captured at each step are adaptively propagated to the image branch through the GCMF module.

\begin{figure}[!t]
\centering
\includegraphics[width=0.98 \linewidth]{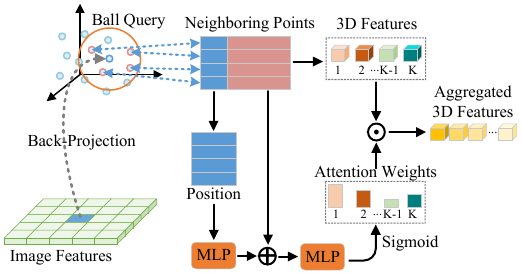}
\caption{Details of the adaptive correlation aggregation strategy.}
\label{aggregate}
\end{figure}

\subsection{Adaptive Correlation Aggregation}
Having extracted the 2D and 3D features, the most critical basis for cross-modal interaction is to appropriately assign 3D features to the corresponding 2D features. To achieve this matching, we can usually establish the correspondence between 2D pixels and 3D points through Eq. (\ref{2DTo3D}). However, this deterministic correspondence relies heavily on accurate depth information. Inaccurate depth values of transparent and specular objects may cause mismatching problem in cross-modal fusion. To this end, we design an adaptive correlation aggregation (ACA) strategy to address this issue, as shown in Fig. \ref{aggregate}. Given a pixel $X_i$ in the 2D feature, we first generate the corresponding point $X_p$ in 3D space based on the predicted depth. Combined with the confidence map $C$ of the depth prediction, we then capture relevant 3D points and features via a radius-adaptive ball query algorithm. The depth prediction with a lower confidence indicates that it is far from the ground truth value, so the ball query radius needs to be enlarged to include more possible relevant point features. Conversely, the prediction with a higher confidence requires a smaller ball query radius to avoid introducing irrelevant point features. Since the confidence $C$ varies in the range $[0, 1]$, we construct a simple linear function to adaptively adjust the radius $r$ of the ball query:

\begin{equation}
    r = C \times r_{min} + (1-C) \times r_{max}.
    \label{radius}
\end{equation}
We select $K$ relevant points from the ball, and we replicate the neighbor points to supplement them when their count falls below the $K$. 

Based on the reference points $X_p$, we combine neighbor point features $F_p^k$ and position information $P_I^k$ to generate learnable attention weights. The position information $P_I^k$ reflects the relationship between the neighbor points $X_p^k$ and the reference point $X_p$:

\begin{equation}
    P_I^k = X_p\oplus X_p^k\oplus (X_p-X_p^k)\oplus \| X_p-X_p^k \|,
    \label{position}
\end{equation}
where $\oplus$ and $\| \cdot \|$ represent the concatenation operation and Euclidean distance respectively. We then use the multi-layer perceptron (MLP) to encode position information and generate attention weights, 

\begin{equation}
    w_f^k = {\rm sigmoid}({\rm MLP}(F_p^k \oplus {\rm MLP}(P_I^k))),
    \label{weights}
\end{equation}
Finally, by weighting the neighbor features using the attention weights, we can get the aggregated relevant 3D features $A_f^p$ of the reference point $X_p$,
\begin{equation}
    A_f^p = \{ w_f^k \odot F_p^k | k = 0, 1,..., K-1\}.
    \label{agg_equ}
\end{equation}
where $\odot$ denotes element-wise multiplication.

\begin{figure}[!t]
\centering
\includegraphics[width=0.9 \linewidth]{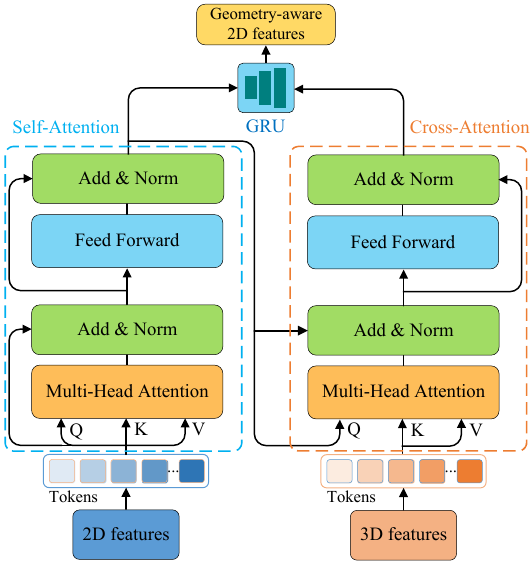}
\caption{The structure of our gated cross-modal fusion module, which consists of self-attention, cross-attention, and gated recurrent unit.}
\label{gcmf}
\end{figure}

\subsection{Gated Cross-Modal Fusion}
After obtaining the aggregated 3D features, we aim to propagate useful 3D features to the image branch and combine them with 2D features to improve the final depth predictions. For this purpose, we develop several gated cross-modal fusion modules to effectively utilize the feature information provided by these two modalities. Specifically, our GCMF module contains three components at each scale: a self-attention, a cross-attention, and a gated recurrent unit (GRU), as presented in Fig. \ref{gcmf}. 

For our self-attention component, we first convert the 2D features into tokens to produce the query vector $Q\in \mathbb{R}^{N\times D}$, key vector $K\in \mathbb{R}^{N\times d}$, and value vector $V\in \mathbb{R}^{N\times d}$, where $N$ and $d$ are the vector length and dimension respectively. We then take $Q$, $K$, $V$ as inputs to capture intra-modality feature correlation and attention.  Formally, a self-attention layer can be simply expressed as:
\begin{equation}
    {\rm Attention}(Q,K,V) = {\rm softmax}(\frac{QK^T}{\sqrt{d}})V.
    \label{self-attention}
\end{equation}
The dot product between $Q$ and $K$ in the vanilla attention layer leads to a computational complexity of $O(N^2)$, which incurs a non-negligible computational time and memory requirement. To alleviate this problem, we adopt the linear attention \cite{linearAtte} with comparable performance to approximate the complex computation and reduce the computational complexity to $O(N)$.

Subsequently, we feed the visual features output from self-attention into our cross-attention component as the query $Q$, and the aggregated 3D features as the key $K$ and value $V$. In this way, each 2D feature can be queried for its unique related 3D features through cross-attention mechanism. Like the self-attention component, we also adopt linear attention here to reduce the amount of computation. Different from simple weighted summation, we finally introduce convolutional GRU \cite{gate} to process the features captured by self-attention and cross-attention components. Convolutional GRU uses the gating mechanism to more effectively retain and transmit multi-modal information useful for depth completion tasks. The features captured by GCMF are sent to the decoder of the hourglass network.

\subsection{Loss Function}
Our method uses the following loss function to jointly optimize the image and point cloud branches:
\begin{equation}
    \mathcal{L} = \mathcal{L}_{I} (D, \hat{D}) + \lambda \mathcal{L}_{P} (P, \hat{P}),
    \label{loss}
\end{equation}
where $\hat{D}$ and $D$ represent the ground-truth and the predicted depth map, $\hat{P}$ and $P$ are the ground-truth and the completed point cloud respectively. The image branch $\mathcal{L}_{I}$ and the point cloud branch $\mathcal{L}_{P}$ are derived from the loss functions used in \cite{li2020multi} and \cite{wen2021pmp}. In the experiment, the loss function weight $\lambda$ is set to 0.01.

\subsection{Object Grasping} \label{object_grasp_section}
By combining our geometry-aware assisted depth completion method with the robotic grasping method, we can evaluate its performance on transparent and specular object grasping tasks. To achieve this, we utilize GraspNet-baseline \cite{fang2020graspnet} as the robotic grasping network, which takes a point cloud as input and predicts 6-DoF grasp poses. Specifically, our method generates an inpainted depth map from the RGB-D input, which is then converted to a point cloud and fed into GraspNet-baseline to predict potential grasp candidates. Finally, the robotic arm performs the grasping operation using the parallel-jaw of the robot arm.

\begin{table}[!t]\large
\centering
\renewcommand{\arraystretch}{1.10}
\resizebox{\linewidth}{!}{
\begin{tabular}{c|cccccc}
\bottomrule
\multirow{2}{*}{Methods} & RMSE $\downarrow$ & REL $\downarrow$ & MAE $\downarrow$ & $\delta_{1.05}$ $\uparrow$ &  $\delta_{1.10}$ $\uparrow$ & $\delta_{1.25}$ $\uparrow$\\ 
\cline{2-7} & \multicolumn{6}{c}{ClearGrasp Syn-known} \\
\hline
ClearGrasp \cite{sajjan2020clear} & 0.034& 0.045& 0.026& 73.53& 92.68& 98.25\\
LIDF-Refine \cite{zhu2021rgb} & 0.012& 0.017& 0.009& 94.79& 98.52 & 99.67\\
DFNet \cite{fang2022transcg} & 0.016& 0.023& 0.013& 89.24& 97.71 & 99.93\\
FDCT \cite{li2023fdct} & 0.014& 0.020& 0.011& 92.21& 97.90 & 99.97\\
TCRNet \cite{zhai2024tcrnet} & \textbf{0.010}& 0.015& \textbf{0.006}& 95.83& 98.74 & 99.75\\
\hline
Ours   & 0.011& \textbf{0.014}& 0.008& \textbf{96.36}& \textbf{98.97}& \textbf{99.99}\\
\hline\hline
 & \multicolumn{6}{c}{ClearGrasp Syn-novel} \\
\hline
ClearGrasp \cite{sajjan2020clear} & 0.037& 0.062& 0.032& 50.27& 84.00& 98.39\\
LIDF-Refine \cite{zhu2021rgb} & 0.028& 0.045& 0.023& 68.62& 89.10 & 99.20\\
DFNet \cite{fang2022transcg} & 0.030& 0.046& 0.024& 66.32& 87.86 & 97.67\\
FDCT \cite{li2023fdct} & 0.026& 0.041& 0.021& 71.09& 92.33 & 99.30\\
TCRNet \cite{zhai2024tcrnet} & 0.023& 0.040& 0.018& 71.33& 90.84 & 99.49\\
\hline
Ours   & \textbf{0.021}& \textbf{0.032}& \textbf{0.016}& \textbf{79.90}& \textbf{94.53}& \textbf{99.75}\\
\toprule
\end{tabular}}
\caption{Quantitative comparison of our GAA-TSO with state-of-the-art methods on the ClearGrasp and OOD datasets. \textbf{Bold} represents the best result.}
\label{cg-ood}
\end{table}

\begin{figure}[!t]
\centering
\includegraphics[width=0.98 \linewidth]{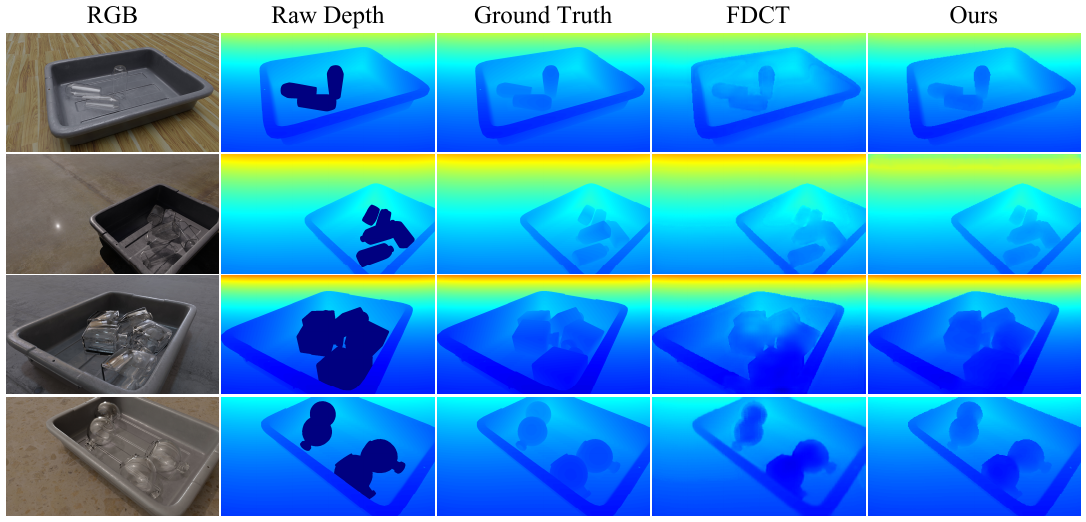}
\caption{Qualitative results on the ClearGrasp dataset. From the first to the fifth column, we show the input RGB, raw depth, ground truth, results of FDCT, and ours.}
\label{cleargrasp_result}
\end{figure}

\section{Experiments}
\subsection{Experimental Setup}
\noindent \textbf{Datasets.} We evaluate our GAA-TSO on four datasets for depth completion of transparent and specular objects, including ClearGrasp \cite{sajjan2020clear}, OOD \cite{zhu2021rgb}, TransCG \cite{fang2022transcg}, and STD \cite{dai2022domain}. ClearGrasp is the first large-scale dataset for transparent objects, which contains 50k synthetic RGB-D images. The OOD dataset contains 60k synthetic RGB-D images of five objects derived from ClearGrasp. TransCG is a real-world transparent object dataset consisting of 58k RGB-D images generated from 130 different scenes. In addition, the STD dataset contains 27k real-world RGB-D images for transparent and specular objects. 

\noindent \textbf{Metrics.} We follow the evaluation metrics used in previous works \cite{sajjan2020clear, zhu2021rgb, fang2022transcg, dai2022domain}, which include root mean squared error (RMSE), absolute relative difference (REL), mean absolute error (MAE), and threshold $\delta$ (set to 1.05, 1.10, and 1.25). Referring to previous studies, these metrics are evaluated only for transparent and specular areas using object masks.

\noindent \textbf{Implementation Details.} By adopting the PyTorch framework, our method is trained and tested on two NVIDIA RTX 3090 GPUs. The Adam optimizer is utilized with an initial learning rate of $10^{-3}$, and input RGB-D images are resized to $320 \times 240$. We train our GAA-TSO from scratch spans 40 epochs with a batch size of 16, and the learning rate is decayed by a factor of 5 at epochs 5, 15, 25, and 35 using a multi-step scheduler. Additionally, the number of relevant point clouds $K$ is set to 16, and the query radius $r_{min}$ and $r_{max}$ are set to 0.05 and 0.1 respectively.

\begin{table}[!t]\large
\centering
\renewcommand{\arraystretch}{1.10}
\resizebox{\linewidth}{!}{
\begin{tabular}{c|cccccc}
\bottomrule
\multirow{2}{*}{Methods} & \multicolumn{6}{c}{Metrics} \\ 
\cline{2-7} & RMSE $\downarrow$ & REL $\downarrow$ & MAE $\downarrow$ & $\delta_{1.05}$ $\uparrow$ &  $\delta_{1.10}$ $\uparrow$ & $\delta_{1.25}$ $\uparrow$ \\
\hline
ClearGrasp \cite{sajjan2020clear} & 0.054& 0.083& 0.037& 50.48& 68.68& 95.28\\
LIDF-Refine \cite{zhu2021rgb} & 0.019& 0.034& 0.015& 78.22& 94.26 & 99.80\\
TranspareNet \cite{xu2021seeing} & 0.026& 0.023& 0.013& 88.45& 96.25 & 99.42\\
DFNet \cite{fang2022transcg} & 0.018& 0.027& 0.012& 83.76& 95.67 & 99.71\\
FDCT \cite{li2023fdct} & 0.015& 0.022& 0.010& 88.18& 97.15 & 99.81\\
TCRNet \cite{zhai2024tcrnet} & 0.017& 0.020& 0.010& 88.96& 96.94 & 99.87\\
\hline
Ours   & \textbf{0.014}& \textbf{0.019}& \textbf{0.009}& \textbf{89.92}& \textbf{98.41}& \textbf{99.96}\\
\toprule
\end{tabular}}
\caption{Quantitative comparison of our GAA-TSO with state-of-the-art methods on the TransCG dataset. \textbf{Bold} represents the best result.}
\label{transcg}
\end{table}

\begin{figure}[!t]
\centering
\includegraphics[width=0.98 \linewidth]{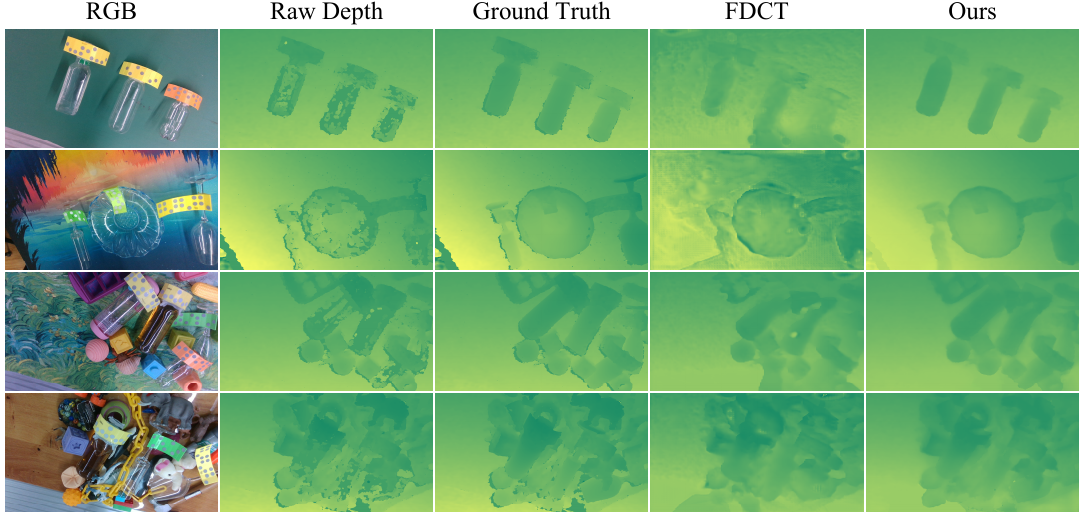}
\caption{Qualitative results on the TransCG dataset. From the first to the fifth column, we show the input RGB, raw depth, ground truth, results of FDCT, and ours.}
\label{transcg_result}
\end{figure}

\subsection{Main Results}
\subsubsection{Evaluation on the ClearGrasp and OOD Datasets}

The OOD dataset consists only of synthetic images, and the objects in the scene are obtained from ClearGrasp. Therefore, to maintain consistency with previous studies, we follow the dataset settings and merge the two datasets for training. For testing, we evaluate our method using synthetic known and synthetic novel test datasets from ClearGrasp. 

Table \ref{cg-ood} presents the quantitative results of our GAA-TSO compared with other recent methods. On the ClearGrasp Syn-known dataset, our approach surpasses others in most metrics. Notably, on the Syn-novel dataset, we show significant advantages across all metrics. Specifically, we outperform TCRNet by 0.008 in REL and 8.57 in $\delta_{1.05}$. These results demonstrate the excellent generalization performance of our approach to previously unseen objects. Figure \ref{cleargrasp_result} provides a visual comparison with the recent method, demonstrating that our method achieves clearer boundary prediction, especially when multiple transparent objects are close together.

\begin{table}[!t]\large
\centering
\renewcommand{\arraystretch}{1.10}
\resizebox{\linewidth}{!}{
\begin{tabular}{c|cccccc}
\bottomrule
\multirow{2}{*}{Methods} & RMSE $\downarrow$ & REL $\downarrow$ & MAE $\downarrow$ & $\delta_{1.05}$ $\uparrow$ &  $\delta_{1.10}$ $\uparrow$ & $\delta_{1.25}$ $\uparrow$\\ 
\cline{2-7} & \multicolumn{6}{c}{STD-CatKnown} \\
\hline
ClearGrasp \cite{sajjan2020clear} & 0.062& 0.067& 0.050& 55.14& 84.46& 92.28\\
LIDF-Refine \cite{zhu2021rgb} & 0.044& 0.049& 0.034& 66.05& 92.13 & 98.07\\
DFNet \cite{fang2022transcg} & 0.035& 0.041& 0.028& 68.02& 94.07 & 99.40\\
SwinDRNet \cite{dai2022domain} & 0.041& 0.036& 0.025& 77.75& 96.24 & 99.55\\
FDCT \cite{li2023fdct} & 0.032& 0.037& 0.026& 72.63& 97.32 & 99.71\\
\hline
Ours   & \textbf{0.029}& \textbf{0.032}& \textbf{0.023}& \textbf{81.13}& \textbf{97.36}& \textbf{99.93}\\
\hline\hline
 & \multicolumn{6}{c}{STD-CatNovel} \\
\hline
ClearGrasp \cite{sajjan2020clear} & 0.088& 0.073& 0.064& 51.72& 79.34& 89.11\\
LIDF-Refine \cite{zhu2021rgb} & 0.053& 0.058& 0.037& 64.52& 85.47 & 93.54\\
DFNet \cite{fang2022transcg} & 0.043& 0.056& 0.030& 66.46& 86.39 & 94.73\\
SwinDRNet \cite{dai2022domain} & 0.050& 0.054& 0.033& 65.77& 85.41 & 94.18\\
FDCT \cite{li2023fdct} & 0.038& 0.051& 0.027& 66.18& 88.63 & 97.45\\
\hline
Ours   & \textbf{0.035}& \textbf{0.047}& \textbf{0.025}& \textbf{74.43}& \textbf{89.72}& \textbf{98.31}\\
\toprule
\end{tabular}}
\caption{Quantitative comparison of our GAA-TSO with state-of-the-art methods on the STD dataset. \textbf{Bold} represents the best result.}
\label{std}
\end{table}

\begin{figure}[!t]
\centering
\includegraphics[width=0.98 \linewidth]{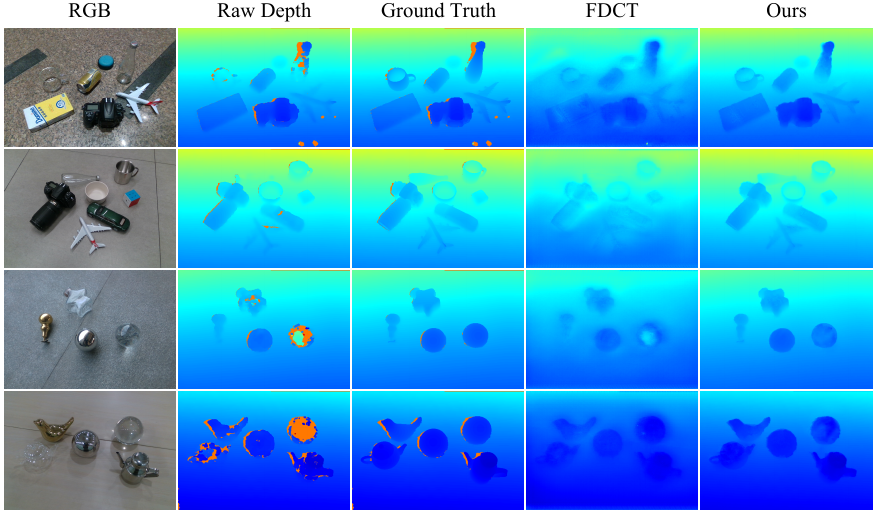}
\caption{Qualitative results on the STD dataset. From the first to the fifth column, we show the input RGB, raw depth, ground truth, results of FDCT, and ours.}
\label{std_result}
\end{figure}

\subsubsection{Evaluation on the TransCG Dataset} 

In this subsection, we use the training and test sets originally divided in TransCG to conduct comparative experiments. We compare our GAA-TSO with recent methods on the real-world TransCG dataset. Quantitative comparison results are reported in Tab. \ref{transcg}. From the experimental results, we can see that our method achieves superior performance in all metrics on the TransCG dataset. Compared with TCRNet, our GAA-TSO has significant improvements in terms of $\delta_{1.05}$ and $\delta_{1.10}$ metrics, which are increased by 0.96 and 1.47 respectively. 

In addition, we also show the qualitative comparisons on the TransCG dataset, as shown in Fig. \ref{transcg_result}. The first two rows are simple scenes, and the last two rows are cluttered scenes. The predictions of our method show promising results compared to FDCT, which contain sharper contours and more details.

\subsubsection{Evaluation on the STD Dataset} 

As for the evaluation of our GAA-TSO on the STD dataset, which contains 30 different scenes (25 STD-CatKnown and 5 STD-CatNovel). Among them, we use 20 STD-CatKnown scenes for training, 5 STD-CatKnown scenes for testing known objects, and 5 STD-CatNovel scenes for testing novel objects. The qualitative results of our GAA-TSO and recent methods are presented in Tab. \ref{std}. These results demonstrate the effectiveness of our GAA-TSO, which outperforms FDCT by 8.25 and 1.09 in terms of $\delta_{1.05}$ and $\delta_{1.10}$ metrics on the STD-CatNovel scenes, respectively.

We also present some visualization results on the STD dataset, as shown in Fig. \ref{std_result}. Benefiting from making full use of geometric information, our method predicts more detailed object structures, while FDCT tends to produce structure-less predictions and blurred boundaries.

\begin{table*}[!t]
\centering
\renewcommand{\arraystretch}{1.10}
\resizebox{0.8\linewidth}{!}{
\begin{tabular}{c|c|c|ccc|cccccc}
\bottomrule
\multirow{2}{*}{Methods} & \multirow{2}{*}{\makecell[c]{Image \\ Branch}} & \multirow{2}{*}{\makecell[c]{Point Cloud\\Branch}} & \multicolumn{3}{c|}{GCMF}  & \multicolumn{6}{c}{Metrics} \\ 
\cline{4-12} & & & $1/4$ & $1/2$ & $1/1$ & RMSE $\downarrow$ & REL $\downarrow$ & MAE $\downarrow$ & $\delta_{1.05}$ $\uparrow$ &  $\delta_{1.10}$ $\uparrow$ & $\delta_{1.25}$ $\uparrow$ \\
\hline
Baseline & \checkmark & & & & & 0.022& 0.031& 0.015& 82.43& 93.51& 99.32 \\
\hline
+3D & \checkmark & \checkmark & & & & 0.021& 0.027& 0.013& 83.82& 94.35& 99.71 \\
\hline
$\rm{GCMF}_{1/4}$ & \checkmark & \checkmark & \checkmark & & & 0.019& 0.026& 0.012& 84.62& 95.05& 99.75 \\
$\rm{GCMF}_{1/2}$ & \checkmark & \checkmark & \checkmark & \checkmark & & 0.017& 0.023& 0.011& 86.99& 97.68& 99.93 \\
Ours & \checkmark & \checkmark & \checkmark & \checkmark & \checkmark & \textbf{0.014}& \textbf{0.019}& \textbf{0.009}& \textbf{89.92}& \textbf{98.41}& \textbf{99.96}\\
\toprule
\end{tabular}}
\caption{Ablation study on the TransCG dataset. \textbf{Bold} represents the best result.}
\label{ablation}
\end{table*}

\begin{table}[!t]\large
\centering
\renewcommand{\arraystretch}{1.10}
\resizebox{\linewidth}{!}{
\begin{tabular}{c|cccccc}
\bottomrule
\multirow{2}{*}{\makecell[c]{Aggregation\\Strategy}} & \multicolumn{6}{c}{Metrics} \\ 
\cline{2-7} & RMSE $\downarrow$ & REL $\downarrow$ & MAE $\downarrow$ & $\delta_{1.05}$ $\uparrow$ &  $\delta_{1.10}$ $\uparrow$ & $\delta_{1.25}$ $\uparrow$ \\
\hline
None  & 0.018& 0.025& 0.012& 86.02& 96.18& 99.27\\
KNN  & 0.016& 0.022& 0.011& 88.29& 97.31& 99.56\\
Ball Query  & 0.015& 0.021& 0.011& 88.41& 97.47& 99.89\\
ACA   & \textbf{0.014}& \textbf{0.019}& \textbf{0.009}& \textbf{89.92}& \textbf{98.41}& \textbf{99.96}\\
\toprule
\end{tabular}}
\caption{Comparison of different neighborhood query strategies. \textbf{Bold} represents the best result.}
\label{ablation_agg}
\end{table}

\begin{table}[!t]
\centering
\renewcommand{\arraystretch}{1.10}
\resizebox{0.9 \linewidth}{!}{
\begin{tabular}{c|cccccc}
\hline
Methods & \#Objects & \#Attempts & GSR & DR \\
\hline
GraspNet & 54& 121 & 44.6 & 41.7\\
DFNet+GraspNet & 67& 113 & 59.1 & 58.3\\
FDCT+GraspNet & 70& 109 & 64.2 & 66.7\\
Ours+GraspNet  & 89& 103 & \textbf{86.4} & \textbf{83.3}\\
\hline
\end{tabular}}
\caption{Results of real robot grasping experiments. \#Objects represents the total number of objects successfully grasped, and \#Attempts indicates the total number of grasping attempts.}
\label{grasping_result}
\end{table}

\subsection{Ablation Studies}

In this subsection, we conduct extensive ablation studies to evaluate the effectiveness of various components in our GAA-TSO.

\noindent \textbf{Effectiveness of point cloud branch.} Here, we use a separate image branch as the Baseline. Based on this, we additionally introduce a point cloud branch to learn the geometric structure information of the scene defined as +3D. By comparing +3D and Baseline in Tab. \ref{ablation}, we can see that the geometric information provided by the point cloud branch helps improve the performance of depth completion. 

\noindent \textbf{Effectiveness of GCMF.} To evaluate the performance of GCMF at different levels, we incrementally incorporate higher-level 3D structures for gated cross-modal fusion. Comparing $\rm{GCMF}_{1/4}$ and +3D in Tab. \ref{ablation}, we find that our proposed GCMF can effectively propagate geometric structure information to the image branch, thereby enriching the spatial perception of 2D features. In addition, high-resolution GCMF can provide more performance improvement than low-resolution ones, because the high-resolution structure provides more geometric details about the object shapes and scene layout.

\noindent \textbf{Effectiveness of ACA.} To evaluate different neighborhood query strategies, we compare the proposed ACA with None, KNN, and Ball Query, where None means not using neighborhood features. As shown in Tab. \ref{ablation_agg}, our ACA significantly enhances the performance, which is attributed to the combination of adaptive query radius and attention weights that enables covering more relevant contexts.

\subsection{Robotic Grasping Experiments}
\begin{figure}[!t]
\centering
\includegraphics[width=0.7 \linewidth]{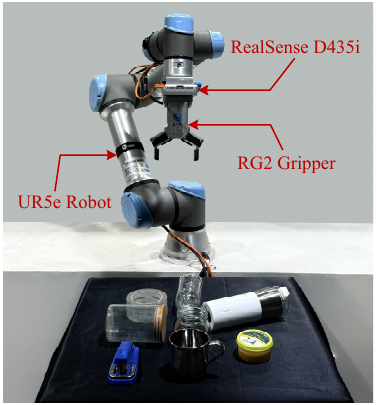}
\caption{Real robot experimental platform for transparent and specular objects grasping.}
\label{real_setup}
\end{figure}

To validate the performance of our approach on downstream robotics tasks, we conduct real-world robotic grasping experiments. Our GAA-TSO incorporated with the object grasping pipeline is introduced in Sec. \ref{object_grasp_section}. We utilize the UR5e robot and OnRobot RG2 gripper as the real experimental platform, which is integrated with the Intel RealSense D435i depth camera as the vision system, as shown in Fig. \ref{real_setup}. 

We conduct 12 rounds of table clearing experiments. In each round, 8 objects are randomly selected, including 3 to 4 transparent and specular objects, and the rest are diffuse objects. The round ends if all objects are removed or grasping an object fails twice. The following metrics are used to evaluate grasping performance: 1) Success Rate: the ratio of successful grasps to the number of attempts, 2) Declutter Rate: the percentage of all objects removed in all rounds. We perform comparison experiments with three baselines: using only GraspNet, using GraspNet with DFNet, and using
GraspNet with FDCT for grasping. Table \ref{grasping_result} presents the results of the robot grasping experiment. Ours+GraspNet outperforms all baselines, demonstrating that our GAA-TSO significantly improves the grasping performance for transparent and specular objects.

\section{Conclusion}\label{conclusion}
In this article, we propose GAA-TSO, a novel geometry-aware assisted depth completion method for transparent and specular objects. To exploit geometric information, we construct a point cloud branch to extract multi-level 3D structural features and design multiple gated cross-modal fusion modules to propagate them to the image branch. In addition, we propose an adaptive correlation aggregation strategy appropriately assign 3D features to corresponding 2D features, which alleviates the projection misalignment problem in cross-modal interaction. Extensive experiments on multiple public datasets demonstrate the effectiveness of our proposed method. We also successfully applied our GAA-TSO to real-world robotic grasping tasks, resulting in a higher grasping performance. In the future, we will explore the use of multi-view perception technology to cope with robotic grasping tasks in more complex scenarios.
\bibliographystyle{IEEEtran}
\bibliography{IEEEabrv, myrefs}

\vfill

\end{document}